%% file: main.tex
\documentclass[journal]{IEEEtran}
\usepackage{blindtext}
\usepackage[pdftex]{graphicx}
\usepackage{amsmath}
\usepackage{amssymb}
\usepackage{amsfonts}
\usepackage{floatrow}
\usepackage{bbm}

\usepackage{multirow} 
\usepackage{rotating}

\usepackage{color}
\usepackage{url}

\definecolor{darkgreen}{rgb}{.0,.3,.3}
\usepackage[colorlinks,citecolor=darkgreen]{hyperref}

\usepackage{enumitem}
\usepackage{logicproof}
\usepackage{subfig}

\usepackage{rotating}
\usepackage{mathtools}

\usepackage[backend=bibtex,maxcitenames=2,natbib=true]{biblatex} 
\definecolor{myred}{rgb}{.8,.0,.0}

\begin{document}

\title{Cats, not CAT scans: a study of dataset similarity in transfer learning for 2D medical image classification}

\author{
    \IEEEauthorblockN{Irma van den Brandt\IEEEauthorrefmark{1}, Floris Fok\IEEEauthorrefmark{1},
    Bas Mulders\IEEEauthorrefmark{1},
    Joaquin Vanschoren\IEEEauthorrefmark{1}
    Veronika Cheplygina\IEEEauthorrefmark{2,1}}\\
    \IEEEauthorblockA{\IEEEauthorrefmark{1}Eindhoven University of Technology, The Netherlands}\\
    \IEEEauthorblockA{\IEEEauthorrefmark{2}IT University of Copenhagen, Denmark}\\
}
\maketitle

\begin{abstract}
Transfer learning is a commonly used strategy for medical image classification, especially via pretraining on source data and fine-tuning on target data. There is currently no consensus on how to choose appropriate source data, and in the literature we can find both evidence of favoring large natural image datasets such as ImageNet, and evidence of favoring more specialized medical datasets. In this paper we perform a systematic study with nine source datasets with natural or medical images, and three target medical datasets, all with 2D images. We find that ImageNet is the source leading to the highest performances, but also that larger datasets are not necessarily better. We also study different definitions of data similarity. We show that common intuitions about similarity may be inaccurate, and therefore not sufficient to predict an appropriate source a priori. Finally, we discuss several steps needed for further research in this field, especially with regard to other types (for example 3D) medical images. Our experiments and pretrained models are available via \url{https://www.github.com/vcheplygina/cats-scans}
\end{abstract}

\section{Introduction}
\input{sec_introduction}

\section{Related Work}
\input{sec_related}

\section{Datasets}\label{ref:datasets}
\input{sec_datasets}

\section{Methods}

\subsection{Transfer learning}
\input{sec_methods_transfer}

\subsection{Meta-learning}
\input{sec_methods_meta}

\section{Results} 

\subsection{Transfer learning}
\input{sec_results_transfer}

\subsection{Meta-learning}
\input{sec_results_meta}

\section{Discussion} 
\input{sec_discussion}

\section*{Acknowledgments}

We thank Colin Nieuwlaat, Felix Schijve and Thijs Kooi for contributing to the discussions about this project. Veronika Cheplygina is supported by the NovoNordisk Foundation (starting package grant) from May 2021, however, the work in this paper was done previously to this award. 

\printbibliography 

\begin{appendix}

\section*{Appendix A - Pretraining settings}\label{sec:app_settings}
\input{sec_appendix}

\end{appendix}

\end{document}

%% file: sec_introduction.tex
In medical image classification, labeled data is often scarce, inviting the use of techniques such as transfer learning~\cite{litjens2017survey,cheplygina2019not,zhou2020review,shang2019what,morid2020scoping}, where the goal is to reuse information across datasets. When training a neural network, transfer can be achieved by first training on a larger, source dataset (for example, natural images such as cats) and then further fine-tuning on a smaller, target dataset (such as a dataset of computed tomography (CT or CAT) scans. This allows the network to reuse features it learned on the source data, thus lowering the amount of target data needed. 

A popular source dataset is ImageNet \citep{imagenet_cvpr09}, although it has been argued whether it is the best strategy for medical target data. For example, \citet{raghu2019transfusion} argue that the dataset properties (i.e. number of classes, level of granularity in classes, size) of ImageNet and a medical target dataset may be too different to allow effective feature reuse. 
More suitable features may be learned from datasets more similar to the medical target dataset \citep{tajbakhsh2019surrogate, teh2020learning}. Some studies have indeed shown that the effectiveness of feature reuse decreased with increasing differences between the source and target task \citep{yosinski2014transferable, tajbakhsh2016convolutional}.  

In a previous study we reviewed papers which compared medical to non-medical source datasets for medical target data~\cite{cheplygina2019cats}. Out of 12 papers, three concluded a medical source was best, three concluded that a non-medical source was best, and two did not find differences. Others did not provide a definite conclusion, but did provide other valuable insights. In general, the papers seemed to agree that source datasets need to be ``large enough'' and ``similar enough'', but exact definitions of these properties were not given. Since each paper used a different set of datasets, it was not possible to extract further conclusions from the study.

Given a new target problem, we would like to identify what the best course of action is without trying all possible source datasets. The goal of the study is therefore two-fold:
\begin{itemize}
    \item Investigate the relationship between transfer learning performance and properties such as dataset size or origin of images, and 
    
    \item Investigate whether a dataset similarity measure, based on such a meta-representation, can be used to predict which source dataset is the most appropriate for a particular target dataset. 
\end{itemize}

We perform transfer learning experiments with nine datasets. We show that the natural image dataset ImageNet, which is the largest, leads to the best performances. However, even small datasets can be valuable for transfer learning. We also examine two definitions of dataset similarity, and how it relates to transfer learning performance. Both definitions show very weak to weak correlations with the performance. However, the most similar datasets appear to have the least effect on performance, which contrasts our results from earlier findings.

%% file: sec_related.tex
\subsection{How can we do transfer learning?}

Transfer learning~\citep{pan2010survey} relies on the idea of transferring information from related, but not the same, learning problems. In supervised classification scenarios we normally assume that the training and test data are from the same \emph{domain} $\mathcal{D} = (\mathcal{X}, p(X)$), and \emph{task} $\mathcal{T} = (\mathcal{Y}, f(\cdot))$, where
$\mathcal{X}$ and $\mathcal{Y}$ are the feature and label spaces, $p(X)$ is the distribution of the feature vectors, and $f$ the mapping of the features to labels. In transfer learning scenarios, we assume that we are dealing either with different domains $\mathcal{D}_S \neq \mathcal{D}_T$ and/or different tasks $\mathcal{T}_S \neq \mathcal{T}_T$. 

Transfer learning can be achieved via different strategies, of which a popular one is to pretrain on the source data, and then use the pretrained network either for extracting off-the-shelf features from the target data, or for further fine-tuning on the target data. In the pretraining/fine-tuning scenario, both the domain and the task can be different. Still, transfer learning can be beneficial, and various studies have looked at how to do this successfully \citep{tajbakhsh2016convolutional,huh2016makes,kornblith2019better,shankar2020evaluating}. Some general findings are for example, that fine-tuning tends to be more beneficial than extracting off-the-shelf features, and that fine-tuning only the last layers is more effective \citet{tajbakhsh2016convolutional}.

\subsection{How are source datasets selected in medical imaging?}

Since the introduction of transfer learning in medical imaging, transferring weights from ImageNet has become popular - \citet{morid2020scoping} review 102 papers which do this. However, there are arguments that the size of images and number of classes might not be suitable for medical imaging, where large (possibly higher-dimensional) images of few classes are typically used~\citep{wong2018building,raghu2019transfusion}.  

Although ImageNet is often used, it is rare to find papers comparing different types of sources. We presented the details of 12 such comparisons in our previous study \cite{cheplygina2019cats}, with a brief summary of the reviewed papers here for completeness:

\begin{itemize}
    \item \citep{schlegl2014unsupervised,menegola2017knowledge,ribeiro2017exploring} conclude that a natural source dataset is better
    \item \citep{shi2018learning,lei2018deeply,wong2018building} conclude that a medical source dataset is better
    \item \cite{cha2017bladder,du2018performance} do not find differences 
    \item \citep{shin2016deep,christodoulidis2017multisource,mormont2018comparison} show that larger source datasets are not necessarily better
    \item \citep{tajbakhsh2016convolutional,zhang2017automatic} suggest similarity to the target is important (but do not compare natural to medical sources)
\end{itemize}

Such studies often mention the concept of dataset similarity, but this definition is often not given or quantified. Similarity can be referred to in terms of the visual image properties \citep{menegola2017knowledge}, feature representations \cite{lei2018deeply}, number and variety of classes \cite{azizpour2015generic}, and other characteristics.

\subsection{How can we predict an appropriate source dataset?}

A way to predict an effective source dataset could be based on meta-learning~\citep{vanschoren2018metalearning}, where meta-data about existing datasets or methods and their performance is available. Such a meta-dataset would allow training a model to predict how well a particular dataset or method would perform in a previously unseen situation. One specific example, and the example that we address, is predicting the suitability of a source dataset. This could be done by only considering the source and target datasets (but not the transfer learning method), which we call ``model-independent''.

Dataset source selection has been studied for various applications.
In an example from natural language processing ~\citep{ruder2017learning}, the authors learn a model-independent measure of dataset similarity, weighting meta-features using examples from the target dataset. They combine similarity measures based on comparing distributions of words or topics with diversity features based on the source data, and show that diversity combines complementary information. 
Another example of dataset source selection is \citep{alvarez2020geometric} where the authors propose a distance measure based on the feature-label distributions (i.e., considering both the domain and the task) of the datasets. They show it is predictive of the transferability of datasets, defined by the relative decrease in error when a source dataset is used. They show this for several applications, although it is worth noting that for natural images, only versions of two related datasets (Tiny ImageNet and CIFAR-10) are examined. 

An example from computer vision is Task2Vec~\citep{achille2019task2vec}, a popular approach to encode tasks (feature-label distributions). The idea behind Task2Vec is to obtain embeddings of classification tasks so that the relationship between the tasks can be analyzed, even if the datasets have different characteristics such as number of classes or image sizes. They investigate both a symmetric and an asymmetric version of the Task2Vec distance and show that the asymmetric version correlates with transferability between tasks. 

Compared to other applied fields, in medical imaging, the application of meta-learning is relatively recent. To date it has been used for predicting segmentation scores in photos of wounds~\cite{campos2016meta}, predicting if two datasets originate from the same source \cite{cheplygina2017exploring}, initializing weights for fine-tuning an algorithm~\cite{hu2018meta} and predicting segmentation scores of segmentation algorithms across 13 datasets \cite{sonsbeek2020predicting}. None of these studies have looked at dataset source selection. 

%% file: sec_datasets.tex
\begin{figure}[!ht]
    \centering
    \includegraphics[width=1\linewidth]{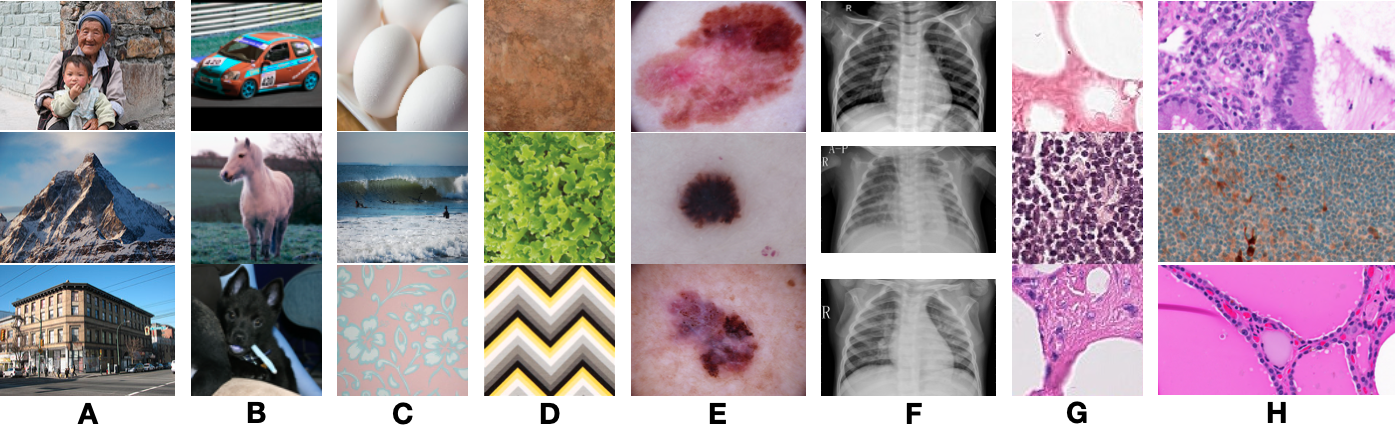}
    \caption[width=1\linewidth]{Three random images of every dataset used in the experiments. From left to right: ImageNet (A), STL-10 (B), STI-10 (C), DTD (D), ISIC2018 (E), Chest X-rays (F), PCam-middle and PCam-small (G), and KimiaPath960 (H). The images shown for ImageNet subsets STL-10 (B) and STI-10 (C) are also included in ImageNet (A).}
    \label{fig:datasets}
\end{figure}

In total, we use nine datasets derived from six data sources in the experiments. Some derived datasets are available online, others are created by sampling from one of the data sources as shown Figure \ref{fig:flowchart}. Table \ref{tab:datasets} provides an overview of the datasets and their properties. In Figure \ref{fig:datasets}, three random examples from every dataset are given. The non-medical datasets and medical datasets PCam-small and KimiaPath 960 are used as source datasets only, the other medical datasets ISIC2018, Chest X-rays and PCam-middle are used as source and target datasets in the transfer learning process. Also, KimiaPath960 is used as source dataset for PCam-middle only. 

\subsubsection{ImageNet} The ImageNet database contains $1,431,167$ natural, colored images of size $256$x$256$ from $1,000$ different classes \citep{imagenet_cvpr09}. On average every class contains $1,000$ images.

\subsubsection{STL-10}
STL-10 is subset of ImageNet consisting of $10$ classes: \textit{airplane, bird, car, cat, deer, dog, horse, monkey, ship} and \textit{truck} \citep{slt10-dataset}. Each class contains $1,300$ $96$x$96$ colored images. 

\subsubsection{Subset Textures ImageNet (STI-10)}
The Subset Textures ImageNet (STI-10) dataset is a subset of ImageNet created for this paper. The idea is to subsample ImageNet to have the same number of images, but not focus on classes depicting textures. The subset consists of $10$ classes representing a textural pattern, which are chosen pseudo-randomly from the ImageNet hierarchy \citep{sti10}. The $10$ classes included are the following: \textit{brick, stone wall, flower, honeycomb, egg, rock, fabric, cloud, chain} and \textit{ocean}. 
On average every class contains $900$ colored images of size $112$x$112$.

\subsubsection{Describable Textures Dataset (DTD)}
The Describable Textures Dataset (DTD) contains images of textural patterns labeled according to $47$ different human-centric attributes inspired by the perceptual characteristics of the patterns \citep{cimpoi14describing}. Every class consists of $120$ colored images with size ranging from $300$x$300$ to $640$x$640$. The textural pattern is displayed on at least $90\%$ of the surface of the image.

\subsubsection{ISIC 2018 - Task 3 (Training set)}
The ISIC 2018 database consists of $10,015$ dermoscopic images of skin lesions collected by the \textit{International Skin Imaging Collaboration (ISIC)} \citep{isic2018, tschandl2018ham10000}. The images are categorized according to seven categories such as melanoma and benign nevus which can be detected from the image. On average every class contains $1,430$ colored images of size $600$x$450$. The maximum class contains $6,705$ images, the minimum class only $115$.
This dataset was released by ISIC as a training dataset for the skin lesion classification task (task 3) in the ISIC Challenge 2018. The test set is not included in this paper since ISIC has never published the associated ground truth labels. 

\subsubsection{Chest X-rays}
The Chest X-rays dataset contains $5,863$ binary-labeled (\textit{Pneumonia/Normal}) chest x-ray images from pediatric patients aged one to five years old collected by the Guangzhou Women and Children's Medical Center \citep{kermany2018identifying}. From the $5,863$ images $1,583$ are labeled normal and $4,273$ are labeled with pneumonia. Image sizes range from $72$x$72$ to $2,916$x$2,583$.

\subsubsection{PatchCamelyon-100,000 (PCam-middle)}
PatchCamelyon (PCam) is a classification dataset consisting of $237,680$ colored images of size $96$x$96$ displaying histopathologic scans of lymph node sections \citep{DBLP:journals/corr/abs-1806-03962}. The images are labeled positive depending on the presence of metastatic tissue, with $163,818$ positive and $163,862$ negative images. A random, stratified subset of $100,000$ images is created to speed up training.

\subsubsection{PatchCamelyon-10,000 (PCam-small)}
Like PCam-middle, PCam-small is a stratified subset of the original PatchCamelyon dataset. The total number of images in this subset is 10\% of PCam-middle, namely $10,000$. 

\subsubsection{KimiaPath960}
The KimiaPath960 classification dataset contains 960 histopathology images of size $308$x$168$ labeled according to 20 different classes \citep{kumar2017comparative}. The twenty classes are based on the image selection process: out of 400 whole slide images of muscle, epithelial and connective tissue 20 images visually representing different pattern types were chosen. After this, 48 regions of interest of the same size were selected and downsampled to $308$x$168$ from all 20 WSI scans.

\input{tables/tab_datasets}

%% file: tables/tab_datasets.tex
\begin{table*}[!ht]
\begin{tabular}{@{}llllll@{}}
\hline

\textbf{Dataset} & \textbf{Origin} & \textbf{Size} & \textbf{Classes} & \textbf{Image Size} & \textbf{Channels}\\ 

\hline

ImageNet \citep{imagenet_cvpr09}  & Non-medical & 1,431,167  & 1,000 & $256$x$256$ & 3\\
STL-10 \citep{slt10-dataset}  & Non-medical & 13,000  & 10 & $96$x$96$ & 3  \\
Subset Textures ImageNet (STI-10) & Non-medical & 8,060 & 10 & $112$x$112$ & 3 \\
Describable Textures Dataset (DTD) \citep{cimpoi14describing} & Non-medical & 5,640  & 47 & [$300$x$300$, $640$x$640$] & 3 \\
ISIC 2018 - Task 3 (Training set) \citep{isic2018, tschandl2018ham10000} & Medical & 10,015 & 7 & $600$x$450$  & 3 \\
Chest X-rays \citep{kermany2018identifying} & Medical & 5,863  & 2 &  [$72$x$72$, $2916$x$2583$]  & 1\\
PatchCamelyon-100,000 (PCam-middle)  & Medical & 100,000 & 2 & $96$x$96$ & 3  \\
PatchCamelyon-10,000 (PCam-small) & Medical & 10,000 & 2 & $96$x$96$ & 3 \\
KimiaPath960 \citep{kumar2017comparative} & Medical & 960 & 20 & $308$x$168$ & 3 \\

\hline
\end{tabular}
\caption{Datasets used in the experiments with properties name, medical or non-medical origin, size, number of classes, image sizes, and number of channels (RGB/Grey-scale). The non-medical datasets ImageNet, STL-10, STI-10, DTD, and the medical datasets PCam-small and KimiaPath960 are used as source dataset only, the other medical datasets ISIC2018, Chest X-rays and PCam-middle are used as source and target dataset.}
\label{tab:datasets}
\end{table*}

%% file: sec_methods_transfer.tex
The first part of our experiments consists of transfer learning experiments. We investigate how different networks, pretrained on the source data with different properties (such as dataset size), perform on the target data.

\begin{figure}[!ht]
    \centering
    \includegraphics[width=1\linewidth]{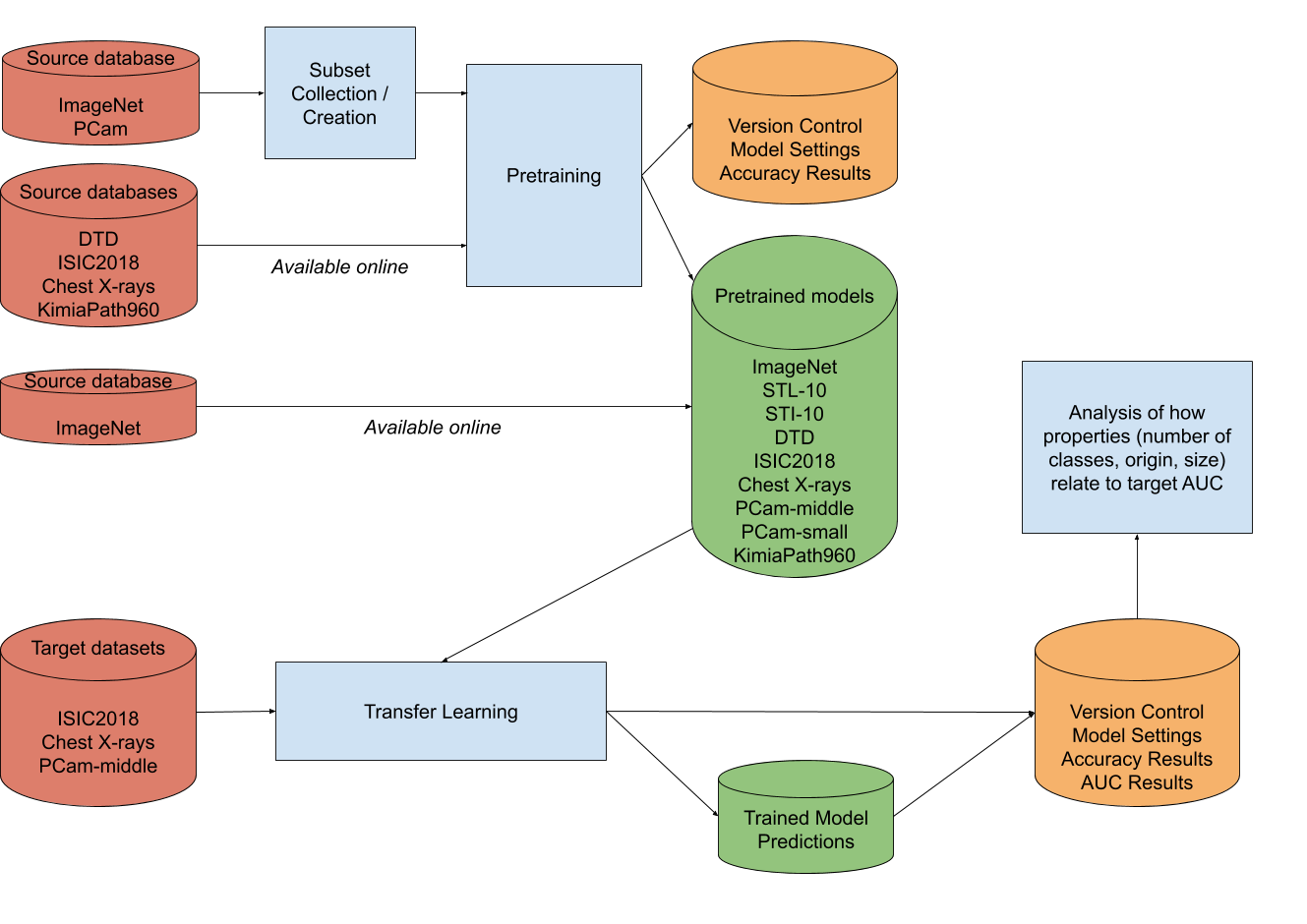}
    \caption{Overview of the experiment setup. Data-related entities are denoted by cylinders, code-entities by rectangles. Red entities are stored on a private server, blue entities on Github, green entities on Open Science Framework (OSF) and orange entities on a Neptune Observer using Sacred.}
    \label{fig:flowchart}
\end{figure}

An overview summarizing the transfer learning experiments different subsections is given in Figure \ref{fig:flowchart}. 

We perform all experiments with the ResNet50-like architecture, as our focus is on the influence of dataset properties on transfer learning, rather than on optimizing performance for a specific dataset. We chose ResNet50 due to their accessibility and ease of optimization~\citep{he2015deep}. We initially also included an EfficientNet-architecture \citep{tan2019efficientnet}, with similar results but slower training, thus we did not continue with these experiments. 

The first step is to pretrain the models. For ImageNet, we used publicly available pretrained models. For the other datasets which are smaller in size, we slightly modify the ResNet50 architecture by replacing the fully-connected layers by a \texttt{GlobalAveragePooling2D}, \texttt{Dropout} and \texttt{Dense} layer for predictions, in order to reduce overfitting~\citep{lin2013network}. We aimed to keep other training settings (dropout rate, data augmentation, image size, batch size, learning rate, training epochs) constant across experiments, but allowed modifications in cases of severe over- or underfitting. The full list of settings can be found in Appendix \ref{sec:app_settings}. These settings were chosen using only the training and validation splits of each dataset.

The second step is to fine-tune the pretrained models with medical target images. For this we adapt the prediction layer of the pretrained model to the target dataset. All other layers are initialized with the resulting weights from pretraining. The whole model is then fine-tuned, as deeper fine-tuning is recommended when the source and target data are from different domains \citep{kornblith2019better,tajbakhsh2016convolutional}. Again, we kept the other settings as constant as possible. 

We evaluate the models using 5-fold cross-validation with $80\%-10\%-10\%$ splits into a training, validation and test sets. We first examined the performance difference of the pretrained models on these sets to diagnose over- or underfitting. We used accuracy for natural images, and AUC (area under the curve) for the medical images. For multi-class problems, we used a one-against-all AUC, weighted by class priors~\citep{scikit-learn}. For evaluating the fine-tuned models on the target datasets, we used 5-fold cross-validation and reported the average AUC-score on the target dataset. 


%% file: sec_methods_meta.tex
In our meta-learning experiments, rather than directly searching for a meta-learning model to predict the most suitable source, we focus on the intermediate step of defining an meta-representation for each dataset, which allows us to measure their similarity. We use two types of meta-features, based on experts, and based on Task2Vec~\citep{achille2019task2vec}.

The \textbf{expert} features are inspired by high-level properties that are sometimes mentioned when image datasets are described. We asked five biomedical engineering students, familiar with transfer learning, to look at nine random images from the dataset, and quantify six properties: natural or medical images, grayscale or color images, objects or textures, the type of objects or textures shown, the diversity of images, and the number of classes. The observers' answers are combined via majority vote, and dataset similarity is calculated via the Hamming distance.

For \textbf{Task2Vec} we use a ResNet50 model pretrained on ImageNet as the probe network. This architecture is chosen for consistency with the transfer learning experiments. For each dataset we then create 100 different subsets using sample weights from the original dataset and the subset index (i.e., 1-100) as random state. The size of each subset is equal to 1\% of the size of the original dataset. The settings such as number of epochs and optimizer used by Task2Vec during training are equal to those in ~\citep{achille2019task2vec}. Running Task2Vec thus creates 100 embeddings per dataset. We calculate the dataset distance as the average cosine distance between five random pairs of embeddings.



To evaluate the quality of these representations, we examine the Spearman correlation between the dataset distance in meta-feature space, and transferability of the datasets. The transferability measures the relative improvement on the target dataset, when a particular source dataset is used. We adapt the measure from~\citep{alvarez2020geometric}, which is based on the classification error, but due to the use of medical target datasets we use the AUC: 

\begin{equation}\label{eq:transferability}
    \mathcal{T}(\mathcal{D}_S \rightarrow \mathcal{D}_T) = \frac{AUC(\mathcal{D}_S \rightarrow \mathcal{D}_T) -  AUC(\mathcal{D}_T) }{AUC(\mathcal{D}_T) } \times 100
\end{equation}

where $AUC(\mathcal{D}_S \rightarrow \mathcal{D}_T)$ is the performance when using transfer learning from source to target, and $AUC(\mathcal{D}_T)$ is the performance of the baseline without transfer learning, i.e. only using the target data.

%% file: sec_results_transfer.tex

\subsubsection{Pretraining}

Table \ref{tab:pretrain_results} shows the results for the source datasets alone, as a baseline to establish that acceptable scores are obtained before moving on to fine-tuning. Note that here single splits are used. 

For the non-medical datasets we evaluate the accuracy. For STL-10, based on natural images, we obtain reasonable accuracy. The accuracy for the texture dataset DTD is rather low - classifying 47 different textural patterns with only 120 images per class is a difficult task. The same holds for STI-10, even though there are now 10 classes with 900 images per class.  

For the medical datasets, AUC is the most relevant, although we report the accuracy for completeness. The AUCs show fairly good discrimination between classes. The difference in PCam-middle and PCam-small shows the effect of reducing the number of images by a factor of 10. Interestingly, the accuracy of both models shows overfitting that is more pronounced than in the other sources.  
\input{tables/tab_pretrain_results}


\subsubsection{Fine-tuning}

\begin{figure}[!ht]
    \centering
    \includegraphics[width=0.95\linewidth]{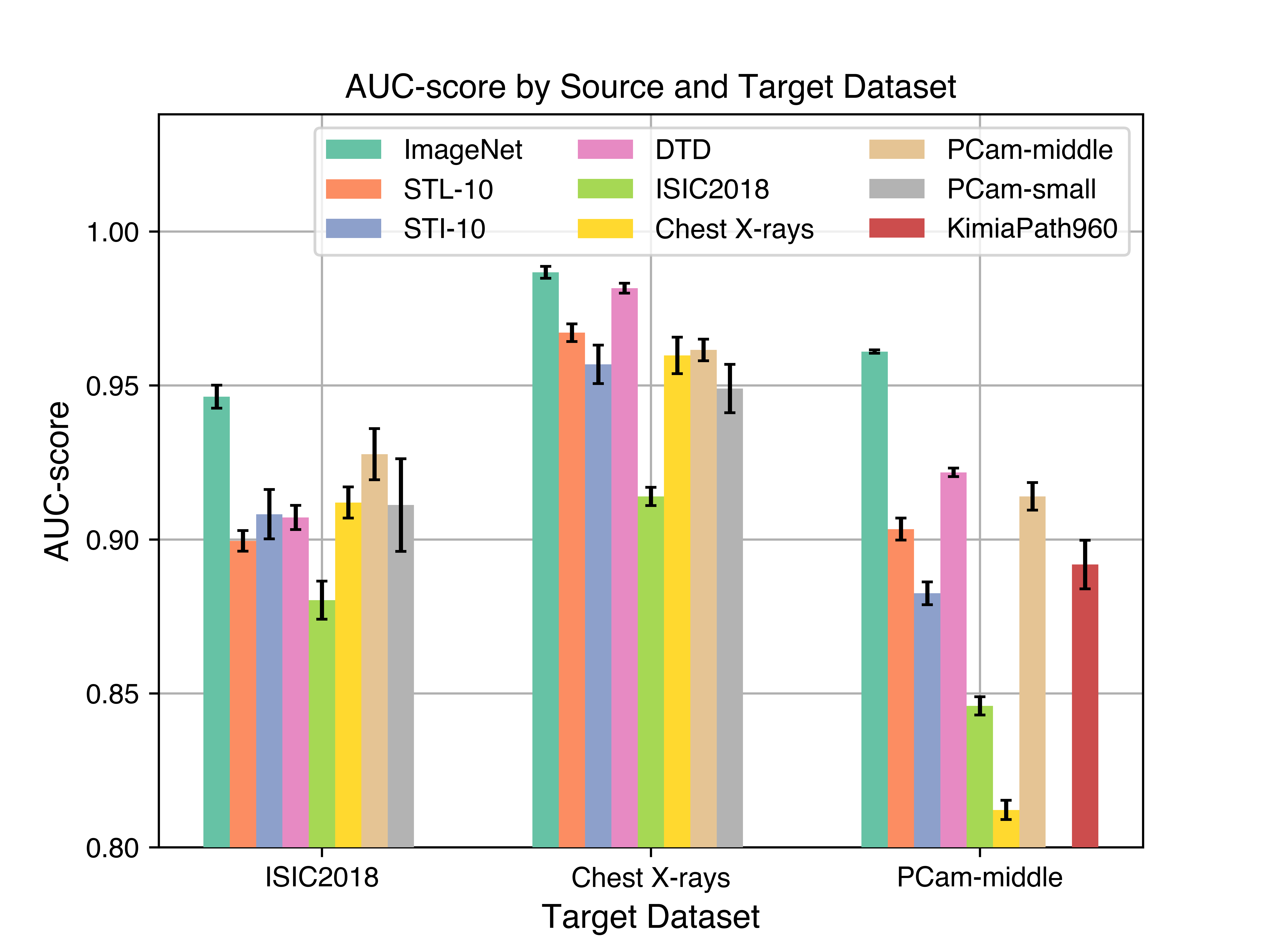}
    \caption{Mean AUC-scores and error bars of five-fold cross validation transfer learning experiments. Source datasets are encoded by a different color bars, target datasets ISIC2018, Chest X-rays and PCam-middle are shown on the x-axis. Using ImageNet weights results in highest scores across all target datasets.}
    \label{fig:AUC_barplot}
\end{figure}

The AUC scores for fine-tuning on the target datasets experiments are given in Figure \ref{fig:AUC_barplot}. For all three targets, ImageNet weights lead to the highest scores. STL-10 and STI-10 both lead to lower performances, although the difference from ImageNet depends on the target dataset. This suggests reducing the number of classes in ImageNet decreases performance, which is consistent with results in \cite{huh2016makes}.

The texture dataset DTD has the second highest scores for ChestX-Ray and PCam-middle, but is not particularly good for ISIC2018. This suggests that the visual importance of textures may be more important for ChestX-Ray and PCam-middle, than for ISIC2018. The high scores on ChestX-Ray and PCam-middle are interesting given the difference in size between DTD and ImageNet. Acquiring more images for all classes in DTD and improving the pretrained model could be worthwhile here. STI-10 performs similarly or worse than DTD in all cases, suggesting that the granularity of DTD might be important. Another potential difference is the resolution of the images, which is lower for STI-10.

Overall, the AUC scores suggest it is not necessarily better to train on medical sources when the target dataset is medical, as we obtained the highest scores with ImageNet. These results do not necessarily align with those of earlier studies. For example, for PCam-middle, \citet{teh2020learning} had better results when training on the full KimiaPath dataset (available only on request and hence not used here) plus the full PCam dataset, suggesting that increasing the size could improve performance. 

On the other hand, our results also show that a large dataset is not a prerequisite for good performance, as shown by training on DTD. This could mean there are opportunities for curating source datasets such that they are a good fit for specific types of target problems, but also more require less training resources due to their smaller size. 

%% file: tables/tab_pretrain_results.tex
\begin{table}[!ht]
\small
\begin{tabular}{@{}lcccr@{}}
\hline
\multicolumn{5}{c}{\textbf{Pretraining Results}} \\
\hline
\textbf{Dataset} & \textbf{Train Acc} & \textbf{Valid Acc} & \textbf{Test Acc} & \textbf{Test AUC} \\ 
\hline
STL-10  & 0.749 & 0.783  & 0.712 &  \\
STI-10 & 0.719  & 0.607  & 0.614 &  \\
DTD & 0.633 & 0.672 & 0.638 & \\ 
ISIC2018 & 0.568 & 0.510 & 0.499 & 0.848 \\
Chest X-rays & 0.810 & 0.804 & 0.790 & 0.912 \\
PCam-middle & 0.801  & 0.709  & 0.710 & 0.834 \\
PCam-small & 0.741 & 0.683 & 0.689 & 0.790 \\
KimiaPath960 & 0.842 & 0.833 & 0.854 & 0.997 \\
\hline
\end{tabular}
\caption{Training, validation and test accuracy scores from pretraining on source datasets. Additionally, for the medical source datasets the AUC-score is given. These are used as a baseline to establish that good enough models are trained on the sources.}
\label{tab:pretrain_results}
\end{table}

%% file: sec_results_meta.tex

\begin{figure}[!ht]
    \centering

    \includegraphics[width=0.95\linewidth]{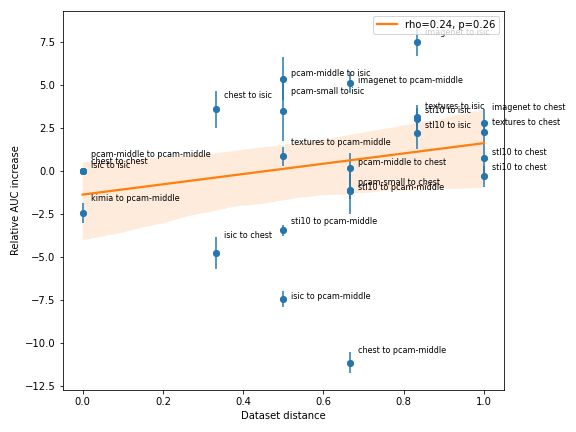}
   
    \includegraphics[width=0.95\linewidth]{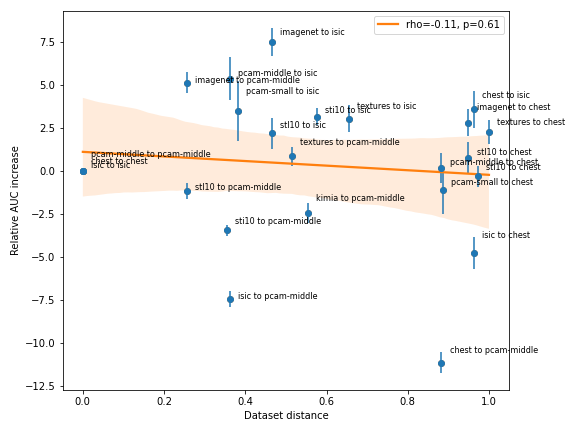}
   
    \caption{Transferability versus dataset distance, defined by (top) experts, and (bottom) task2vec.}
    \label{fig:transferability}
\end{figure}

We show the relationship between the dataset distances and the transferability (defined in Eq. \ref{eq:transferability}) in Fig. \ref{fig:transferability}. The expert assessments show a weak correlation ($\rho=0.24$), and the task2vec representation shows a very weak negative correlation ($\rho=-0.11$). Neither correlation is significant, if we use 0.05 as the significance threshold.  

There are, however, some trends we can observe in the plots. A general trend is that smaller dataset distances do not affect the transferability a lot. Although research often refers to ``similar'' datasets being effective, intuitively the most similar dataset is the target dataset itself, which would not add new information to the learning problem.

On the other hand, larger dataset distances can impact the transferability both positively and negatively. While ImageNet has a higher distance to most datasets, presumably its size and diversity ensure positive transferability scores for most datasets. On the other hand, Chest-XRay might be too different from the other datasets we used, possibly because of grayscale rather than RGB images, and sometimes even leads to negative transferability scores.

%% file: sec_discussion.tex
To summarize, our results show that:

\begin{itemize}
    \item ImageNet weights lead to the best AUC scores for all three medical targets
    \item The dataset size and number of classes are important, but there are exceptions, such as with the texture dataset DTD
    \item Medical sources do not necessarily outperform non-medical sources
    \item ``Similar'' datasets often do not improve training, ``dissimilar'' datasets can affect training both positively and negatively
\end{itemize}

There are a number of important considerations when interpreting these results, which we discuss below.

\subsection{Transfer learning}
We estimated transfer learning performance using a modified ResNet50 model, and allowed for minor changes to the parameters to prevent severe over- or under-fitting. Despite allowing for this, a few models did not converge during training. Since we tried to avoid exhaustively optimizing for each dataset, this is a feature rather than a bug. We also had to make some concessions (such as resizing the images) due to computational resource constraints. On the other hand, for ImageNet we used publicly available weights, where more resources have been allocated to optimization. Therefore, our results might not necessarily reflect the performances typically found in the literature, when less datasets are examined at a time.

We only report results for a ResNet50-like architecture, however in early experiments, we obtained similar results with an EfficientNet, which was less computationally efficient, which is why we did not continue with the experiments. Despite ResNet's suitability for transfer learning, for smaller datasets with fewer classes like the medical datasets, over-parameterization is a serious issue with this model \citep{raghu2019transfusion}, which is consistent with our results. Based on these findings, it would be worthwhile to repeat the experiment using smaller, more lightweight models \citep{DBLP:journals/corr/LiKDSG16}. 

\subsection{Meta-Learning}
We used two different ways to quantify the similarity of datasets - based on a data-driven Task2Vec representation, and based on human assessments of high-level features of the datasets, such as the type of images. The Task2Vec representation may not be the best choice, because it is based on an ImageNet-trained network, possibly biasing the results. The expert representation is based on selected properties, and would be different if other factors were included.   

Finally, our experiments encompass a limited selection of (types) of datasets. In particular, we chose to focus on 2D datasets due to the ease of adaptation of training the models, and all datasets except Chest-XRay consisted of RGB images. As such, the transferability plots likely only show a part of the whole picture. Including other modalities, particularly 3D images such as computed tomography (CT) or magnetic resonance (MR) images, would increase the range of dataset distances, and possibly also transferability scores. 

Even with an increased range of dataset distances and transferability scores, we cannot expect generalization to all datasets and tasks. Locally, however, we expect there to be useful patterns, that can reduce the computational resources needed when selecting a source dataset for a previously unseen target medical imaging problem.

\subsection{Future directions}

Next to studying other transfer learning and meta-learning methods, we expect that the most informative experiments will involve using other datasets. These include other medical imaging modalities such as CT or MR, but also including modified versions of the source datasets. It would be valuable to have more fine-grained variations in various dataset properties, such as the size. 

There is also a connection between data augmentation and transfer learning that is worth investigating. Introducing more diversity into the training data, as data augmentation does, can be beneficial, but can also be detrimental if the changes are too large. As such it would be interesting to interpolate between data augmentation and transfer from a different source, for example by creating ``mix-and-match'' source datasets. 

These experiments are resource-intensive, especially if larger datasets are used. We would therefore emphasize the importance of researchers sharing their pretrained models. This could improve performance across datasets, but also helps with reducing the overall carbon footprint of our field \cite{anthony2020carbontracker} and prevents its de-democratization \cite{ahmed2020democratization}. Ideally we envision a platform like OpenML\footnote{\url{https://www.openml.org}}, where sharing of datasets,  experiments and results would be possible. Until then we encourage the use of persistent repositories, such as Github and OSF. Our own code and models can be accessed via \url{https://www.github.com/vcheplygina/cats-scans}.

\section{Conclusions}

We studied transfer learning from natural and medical 2D image datasets, and investigated dataset similarity in relation to the performance after transfer. We show that ImageNet leads to the best performances after transfer, but our results also suggest that even small datasets can be valuable for transfer. We also show that our intuition about dataset similarity is not necessarily predictive of transfer learning performance. We expect that further insights can be gained by studying more diverse datasets and connections with for example data augmentation.

%% file: sec_appendix.tex
\subsection{Pretraining}

\begin{table*}[!ht]
\begin{tabular}{@{}llllll@{}}
\hline
\textbf{Source Dataset} & \textbf{Image size} & \textbf{Batch Size} & \textbf{Learning Rate} & \textbf{Learning Rate Scheduler} & \textbf{Epochs} \\ 
\hline
STL-10  & $96$x$96$ & 128 & 1.0$e-$3 & Yes, after 30 epochs & 50 \\
STI-10 & $112$x$112$ & 128 & 1.0$e-$4 & Yes, after 50 epochs & 70 \\
DTD & $300$x$300$ & 12 & 1.0$e-$3 & No & 60 \\ 
ISIC2018 & $112$x$112$ & 128 & 1.0$e-$5 & No & 70\\
Chest X-rays & $112$x$112$ & 128 & 1.0$e-$6 & No & 70\\
PCam-middle & $96$x$96$ & 128 & 1.0$e-$6 & No & 50 \\
PCam-small & $96$x$96$ & 128 & 1.0$e-$6 & No & 70 \\
KimiaPath960 & $308$x$168$ & 12 & 1.0$e-$5 & No & 100 \\
\hline
\end{tabular}
\caption{Image sizes and compilation settings for all source datasets used during pretraining. Image sizes are either based on the minimal image size in the dataset, or to avoid long training times or unnecessary resizing. In case of larger image sizes the batch size is minimal to avoid memory issues. Compilation settings are chosen such that the model converges and overfitting is minimized.}
\label{tab:pretrain_settings}
\end{table*}

Generators feed the images in mini-batches to the model, the size of the mini-batch given a specific source dataset is given in Table \ref{tab:pretrain_settings}. For DTD and KimiaPath960 a small batch size of 12 is used due to the large image sizes, for all other datasets a large batch size of 128 is used. 

The different image sizes and model compilation settings for all source datasets are given in Table \ref{tab:pretrain_settings}. In most cases the image size is based on the minimal image size of the dataset (Table \ref{tab:datasets}). For ISIC2018 a smaller size is chosen to speed up training, for Chest X-rays a larger size to avoid unnecessary resizing. 

The learning rate, used as argument for the \textit{Adam}-optimizer, and number of epochs are set to minimize overfitting and (almost) enable model convergence. In some cases a learning rate scheduler, that multiplies the learning rate by $0.5$ after $x$ amount of epochs, is used to slow down learning and avoid overfitting. Potential class imbalance is minimized by passing class weights to the model. 

\subsection{Fine-tuning}

The images are fed in mini-batches of size 128 to the model after being augmented with the same small augmentations as used during pretraining to potentially improve transfer performance results \citep{morid2020scoping}.

All images are resized to $112$x$112$ for experiments where ISIC2018 or Chest X-rays is used as target dataset. For experiments involving PCam-middle as target dataset the images are not resized, their original size of $96$x$96$ remains unchanged. The \textit{Adam}-optimizer is used with learning rates varying between 1.0$e-$4 and 1.0$e-$6 for different target datasets. In experiments involving target datasets ISIC2018 or Chest X-rays the number of epochs is set to 50, for PCam-middle the number of epochs is 20 to avoid long training times. The learning rates and number of epochs are chosen such that (severe) overfitting is avoided and the model (almost) converges. Class weights are provided to the model to minimize potential class imbalance. 